\documentclass[conference]{IEEEtran}
\IEEEoverridecommandlockouts

\usepackage{cite}
\usepackage{amsmath,amssymb,amsfonts}
\usepackage{graphicx}
\usepackage{booktabs}
\usepackage{textcomp}
\usepackage[T1]{fontenc}
\usepackage[utf8]{inputenc}
\usepackage{url}
\usepackage{microtype}
\usepackage[hidelinks,breaklinks]{hyperref}

\graphicspath{{figs/}}

\begin{document}

\title{Feasibility Ceilings Before Learning: Sizing and Timing Bounds
for Drone-Mesh Interception}

\author{\IEEEauthorblockN{Dmytro Babarytskyi}
\IEEEauthorblockA{\textit{Kharkiv National University of Radio Electronics}\\
Kharkiv, Ukraine\\
dmytro.babarytskyi@nure.ua\\
babaritskiydmytro@gmail.com\\
ORCID: 0009-0009-3614-9601}
}

\maketitle

\begin{abstract}
A mesh of interceptors defending a fixed site must spread over the predicted
arrival point of an approaching object. This is a recurring task in multi-agent
robotics: $N$ agents, each covering a disc of radius $r$, spread over a
probability density so that a point drawn from it falls within some agent's
disc. We give closed forms for the quantities that bound performance before any
policy is chosen, and validate each by Monte Carlo. The uncertainty of a point
predicted by extrapolating a noisy linear track follows from least-squares
variance propagation plus a delta-method term for the uncertain arrival time,
agreeing with simulation to within $1.8\%$. The coverage probability of a
packed mesh admits a closed form accurate to $0.007$ absolute, with a
volumetric analogue. A naive area ratio misses a ceiling: while agents stay at
least $2r$ apart, no mesh exceeds $90.69\%$ coverage in the plane or $74.05\%$
in a volume. This ceiling belongs to the separation constraint, not to the
problem --- permitting overlap substitutes Kershner's thinnest-covering density
and removes it, the better regime above a crossover at $p=0.80$. Hexagonal
packing is optimal among five arrangements; the centroidal-Voronoi
configuration coverage control supplies reaches only half its coverage at small
budgets. The Gaussian assumption is conservative under heavy tails and
anisotropy but fragile under bias: a one-sigma aim-point offset costs $14$
points of coverage. A repositioning team must also choose a commit time, and
the admissible window can be empty. Against the applicable ceiling, analytic
placement attains $85\%$ of what the configuration permits while centralised
PPO averages $15\%$ over three trainer seeds, with a paired bootstrap against a
null policy containing zero at every seed. We argue such ceilings should be
computed before any learning experiment and quoted alongside every result.
\end{abstract}


\section{Introduction}\label{sec:intro}

A fixed site is approached by an airborne object on a ballistic or
near-ballistic course. A sensor tracks the object and predicts where it will
cross a defensive plane standing off from the site, with an error that shrinks
as the track lengthens. A mesh of small interceptors, each able to engage
whatever comes within a radius $r$ of it, must spread itself across that
predicted crossing point and be on station before the object arrives. Three
questions decide whether the concept is viable at all, and none of them is
about control: how many interceptors the prediction's uncertainty demands, how
early the mesh must commit to a position, and whether any commit time satisfies
both the sharpness of the prediction and the distance the vehicles must fly.
All three can be answered from geometry, before a controller exists.

We strip the application to that geometry. Consider a team of $N$ agents
constrained to a plane, each able to cover a disc of radius $r$ centred on
itself. A point of interest will arrive somewhere in that plane at a known
future time; the agents hold a noisy prediction of where.
The team must distribute itself so that the arriving point falls inside some
agent's disc. This is a coverage problem in the sense of Cort\'es
\textit{et al.}~\cite{cortes2004}, specialised in two ways that turn out to
matter a great deal. The density is a prediction with a quantified and
time-varying error rather than a static importance map. And success is binary
and instantaneous --- a point either lands within $r$ of some agent or it does
not --- rather than an integrated sensing cost.

Tasks of this shape invite a multi-agent reinforcement-learning treatment, and
the recipe is easy to follow: define an environment, shape a reward, train a
policy, report a success rate. The difficulty is that a bare success rate on a
hand-designed task carries almost no information. Whether $16\%$ is a failure
or a near-optimal result depends on what the configuration permits, and that
quantity is rarely computed. Henderson \textit{et al.}~\cite{henderson2018},
Colas \textit{et al.}~\cite{colas2018} and Agarwal
\textit{et al.}~\cite{agarwal2021} have each documented how fragile deep
reinforcement-learning comparisons are to precisely this kind of missing
context --- unreported baselines, too few seeds, and point estimates quoted
without dispersion.

This paper supplies the missing context for the coverage task above, in closed
form, and then validates the assumptions the closed form rests on. We
contribute:

\begin{enumerate}
\item the uncertainty $\sigma$ of the predicted arrival point, propagated from
      per-fix sensor noise through a least-squares track fit, with a
      delta-method term that a common heuristic omits at a cost of up to
      $9.4\times$ (Section~\ref{sec:prediction});
\item the coverage probability attainable by $N$ agents at given $\sigma$ and
      $r$, in the plane and in a volume, including a ceiling no mesh can exceed
      at any size (Section~\ref{sec:coverage});
\item an empirical check that hexagonal packing is the right arrangement, and a
      quantification of how far the standard coverage-control answer falls
      short of it on this objective (Section~\ref{sec:strategies});
\item a robustness study of the Gaussian assumption, separating departures that
      make the law conservative from the one that breaks it
      (Section~\ref{sec:robustness});
\item the admissible window of commit times for a team that must physically
      reposition, which can be empty (Section~\ref{sec:commit}).
\end{enumerate}

Section~\ref{sec:learning} reports what happened when a learned placement
policy was trained on this task, across three trainer seeds, and argues that
the outcome is better explained by these bounds than by any property of the
learner. Section~\ref{sec:limits} states what the analysis omits.

Each individual result applies standard material --- ordinary least squares,
the delta method, densest packing of equal spheres. The claim is that
computing them \emph{first}, and reporting them \emph{alongside} learned
results, changes what those results mean, and that this is not current
practice.

\section{Related Work}\label{sec:related}

\subsection{Coverage control}

The canonical formulation distributes agents over a density to minimise an
integrated sensing cost, and is solved by Lloyd-type descent to a centroidal
Voronoi configuration~\cite{cortes2004,du1999}. The framework has been extended
to agents that learn the density online~\cite{schwager2009} and developed into
a general theory of distributed robotic networks~\cite{bullo2009}. More
recently, graph neural networks have been used to let agents exploit multi-hop
information and beat the purely local classical controller~\cite{gosrich2022}.

The objective in all of this work is continuous: a weighted integral of the
distance from each point to its nearest agent. Our objective is the probability
that a single draw lands within a fixed radius of some agent. The two are not
equivalent, and Section~\ref{sec:strategies} measures the gap: at small
budgets, the centroidal Voronoi configuration attains barely half the coverage
of a packed one, because minimising mean squared distance rewards spreading
into the tails while binary containment does not.

\subsection{Packing bounds}

That the densest packing of equal circles in the plane has density
$\pi/\sqrt{12}$ is classical, due to Thue and developed by Fejes
T\'oth~\cite{fejestoth1953,conway1999}; the volumetric analogue $\pi/\sqrt{18}$
is the Kepler conjecture, proved by Hales~\cite{hales2005} and later verified
formally by machine~\cite{hales2017}. We use these as the ceiling on what any
non-overlapping arrangement of covering discs can achieve --- a qualification
that matters, and which Section~\ref{sec:regimes} makes explicit by showing
what happens when the non-overlap constraint is dropped. To our knowledge
this ceiling is not usually stated in the multi-agent coverage literature,
where it is natural to write down an area ratio that tacitly assumes perfect
tiling and therefore permits coverage probabilities that are unattainable at
any team size.

\subsection{Track prediction and manoeuvre models}

Propagating measurement noise through a least-squares trajectory fit is
standard estimation practice~\cite{barshalom2001}, as is the recursive
alternative~\cite{kalman1960}. We use the batch closed form rather than a
filter because the quantity wanted is not a running state estimate but the
error covariance of an extrapolation to a fixed plane, and because a closed
form keeps the resulting sizing law analytic end to end. The delta method used
for the uncertain arrival time is likewise textbook~\cite{oehlert1992}.

The straight-line assumption is the weak point, and the tracking literature has
long known it. Singer~\cite{singer1970} models target acceleration as a
correlated random process precisely because constant-velocity fits degrade
badly under manoeuvre, and Li and Jilkov~\cite{li2003} survey several decades
of successors. Section~\ref{sec:commit} uses a deliberately crude version of
this idea --- an independent per-axis acceleration with fixed standard
deviation --- and finds that it dominates the sensor term by two orders of
magnitude at realistic lead times. A Singer-type correlated model would refine
the constant; it would not change which term binds.

\subsection{Assignment}

Allocating agents to positions at minimum total travel is a linear assignment
problem, solved exactly by the Hungarian method~\cite{kuhn1955} in $O(N^{3})$.
Distributed alternatives exist where centralised computation is
unavailable~\cite{bertsekas1988,zavlanos2008,choi2009}. This matters below
because assignment is one of two classical sub-problems that an end-to-end
policy is implicitly asked to rediscover, and the only one with an exact
polynomial solution.

\subsection{Interception and pursuit}

Coverage of a predicted arrival point is adjacent to pursuit-evasion, surveyed
for mobile robotics by Chung \textit{et al.}~\cite{chung2011}. The distinction
is that the team here does not pursue: it positions itself ahead of the
arrival and lets the geometry do the rest, which converts a speed problem into
a placement problem. That is also why packing, rather than pursuit strategy, is
the governing mathematics.

\subsection{Multi-agent reinforcement learning and its evaluation}

PPO~\cite{schulman2017} remains a common default for continuous multi-agent
control, and is a strong baseline in cooperative settings when implemented
carefully~\cite{yu2022}; we use the Stable-Baselines3
implementation~\cite{raffin2021}. Permutation-equivariant
architectures~\cite{battaglia2018,liu2020,vanderpol2022} are the standard
remedy for policies whose action space treats interchangeable agents as
unrelated output dimensions. Our negative result in
Section~\ref{sec:learning} concerns an unstructured centralised baseline, and
we are explicit that it does not settle the structured case.

Reward-specification pathologies are well understood in
principle~\cite{ng1999,skalse2022,sutton2018}; what we add is a concrete
instance in which two of them produced the same observable symptom as a genuine
learning failure, and were diagnosed only after the fact. On evaluation, we
follow the recommendations of \cite{henderson2018,colas2018,agarwal2021} in
reporting multiple seeds with interval estimates rather than point estimates,
using the percentile bootstrap~\cite{efron1993}.

\section{Problem Formulation}\label{sec:problem}

A point of interest travels on a straight line and will cross a fixed plane
$x=x_{p}$ at an unknown in-plane location $\mathbf{c}\in\mathbb{R}^{2}$. A
sensor supplies $n$ position fixes at times $t_{1},\dots,t_{n}$, each corrupted
by independent Gaussian noise of per-axis standard deviation $\sigma_{s}$.

$N$ agents occupy the plane. Each covers a disc of radius $r$; $r$ is treated
throughout as a fixed input parameter of the platform, not a design variable.
Agents move with bounded speed $v_{\max}$ and bounded acceleration $a$, and
must be stationary on station when the crossing occurs. Success is the event
that $\mathbf{c}$ lies within $r$ of at least one agent, and the objective is
its probability,
\begin{equation}
p \;=\; \Pr\Big[\min_{i\in\{1,\dots,N\}}\|\mathbf{c}-\mathbf{q}_{i}\|\le r\Big],
\label{eq:objective}
\end{equation}
over agent positions $\mathbf{q}_{i}$ subject to a minimum separation $s$.

The parameter values carried through the paper belong to the setting of
Section~\ref{sec:intro}: $\sigma_{s}=4$ m per axis is representative of a radar
track on a small airborne object at the ranges involved, $v_{\max}=33$ m/s and
$a=27$ m/s$^{2}$ are of the order of a small multirotor interceptor, and the
manoeuvre magnitudes swept in Section~\ref{sec:commit} run from $0.1g$, a
nearly ballistic object, to $3g$, one that is actively evading.

Two modelling choices need stating, because the results depend on them and
neither is forced. First, $r$ is a \emph{coverage} radius, a property of
whatever the agent senses or affects, whereas $s$ is a \emph{physical}
constraint on how close two agents may fly. These are distinct quantities and
need not coincide. Throughout Sections~\ref{sec:coverage}
to~\ref{sec:commit} we assume $s\ge2r$, so that coverage discs are disjoint and
the arrangement is a packing; Section~\ref{sec:regimes} treats the alternative
and shows it changes the conclusions materially. In the configuration of
Section~\ref{sec:learning}, $s=2.5$ m against $2r=6$ m, so the packing regime
there is a choice of arrangement rather than a constraint of the platform.

Second, $N$ is treated as a continuous quantity wherever it is the output of a
sizing formula. A requirement of ``$22.1$ agents'' means that $22$ are
insufficient and $23$ suffice; we report the continuous value because rounding
discards the margin, which matters when comparing designs whose requirements
differ by less than one agent.

We write $\sigma$ for the per-axis standard deviation of the predicted crossing
point and $R=\sigma\sqrt{2}$ for its RMS radial error. Packing densities are
$\eta_{2}=\pi/\sqrt{12}\approx0.9069$ in the plane and
$\eta_{3}=\pi/\sqrt{18}\approx0.7405$ in a volume. $T$ denotes the commit time,
the lead with which the team fixes its aim point, and $C$ the half-width of the
corridor within which the crossing may fall.

All experiments import the prediction, lattice and environment code from the
accompanying prototype rather than reimplementing it, so that a validation run
tests the implementation that actually runs. Numerics use
NumPy~\cite{harris2020} and SciPy~\cite{virtanen2020}.

\section{Prediction Uncertainty}\label{sec:prediction}

\subsection{Closed form}

Fit per-axis ordinary least-squares lines $p_{k}(t)=a_{k}+b_{k}t$ for
$k\in\{x,y,z\}$ and solve $p_{x}(t)=x_{p}$ for the crossing time $t_{c}$. With
$\bar{t}$ the mean fix time and $S=\sum_{i}(t_{i}-\bar{t})^{2}$, the standard
OLS results are
\begin{equation}
\begin{aligned}
&\mathrm{Var}(b)=\frac{\sigma_{s}^{2}}{S},\quad
\mathrm{Var}(a)=\sigma_{s}^{2}\!\left(\frac{1}{n}+\frac{\bar{t}^{2}}{S}\right),\\
&\mathrm{Cov}(a,b)=-\frac{\sigma_{s}^{2}\bar{t}}{S}.
\end{aligned}
\label{eq:ols}
\end{equation}
The fitted value at an arbitrary time $\tau$ therefore has variance
\begin{equation}
\mathrm{Var}\big(a+b\tau\big)=\mathrm{Var}(a)+\tau^{2}\mathrm{Var}(b)
 +2\tau\,\mathrm{Cov}(a,b).
\label{eq:varat}
\end{equation}
The crossing time is itself an estimate. Applying the delta
method~\cite{oehlert1992} to $t_{c}=(x_{p}-a_{x})/b_{x}$, with
$\partial t_{c}/\partial a_{x}=-1/b_{x}$ and
$\partial t_{c}/\partial b_{x}=-t_{c}/b_{x}$,
\begin{equation}
\mathrm{Var}(t_{c})=\frac{\mathrm{Var}(a_{x})}{b_{x}^{2}}
 +\frac{t_{c}^{2}}{b_{x}^{2}}\mathrm{Var}(b_{x})
 +\frac{2t_{c}}{b_{x}^{2}}\mathrm{Cov}(a_{x},b_{x}).
\label{eq:vartc}
\end{equation}
Each in-plane axis then carries its own fit variance at $t=t_{c}$ plus a term
from $t_{c}$ being uncertain, since to first order the fitted value shifts by
$b_{k}\,\delta t_{c}$. Combining,
\begin{equation}
R^{2}=\sum_{k\in\{y,z\}}\Big[\mathrm{Var}\big(a_{k}+b_{k}t_{c}\big)
      + b_{k}^{2}\,\mathrm{Var}(t_{c})\Big].
\label{eq:R}
\end{equation}

The second term is easy to omit and expensive to omit. Error in the fitted
closing velocity does not merely blur the crossing \emph{time}: it is
multiplied by $t_{c}$ before it reaches the in-plane prediction, so a small
slope error compounds into a large position error at long lead. Two practical
notes follow. The term vanishes identically for a point travelling exactly
along the $x$ axis, so a validation scenario built without lateral velocity
will not exercise it. And because the sensor noise is identical and independent
across axes in our setting, there is no cross-term between the $x$ fit and the
$y,z$ fits to add.

\subsection{Validation}

Table~\ref{tab:e1} reports \eqref{eq:R} against Monte Carlo over noisy
re-draws of a fixed true trajectory, on a $4\times4$ grid. The grid is
parameterised by $(n,t_{c})$ rather than by $(n,\text{standoff})$: holding the
standoff fixed while adding fixes consumes the whole approach and leaves
nothing to extrapolate, which confounds the two effects under test. The
standoff is solved for instead, so every cell is a valid geometry.

Agreement is within $1.8\%$ in the worst cell. Calibration --- the fraction of
trials whose true crossing point falls inside the radius the predictor reported
--- holds between $61.7\%$ and $64.3\%$ against the $63.2\%$ nominal for a
one-sigma radial bound on a two-dimensional Gaussian. This is the stronger of
the two checks: agreement in magnitude could be arranged by a lucky constant,
whereas calibration across sixteen geometries tests the shape of the
distribution as well.

\begin{figure*}[t]
\centering
\includegraphics[width=0.92\textwidth]{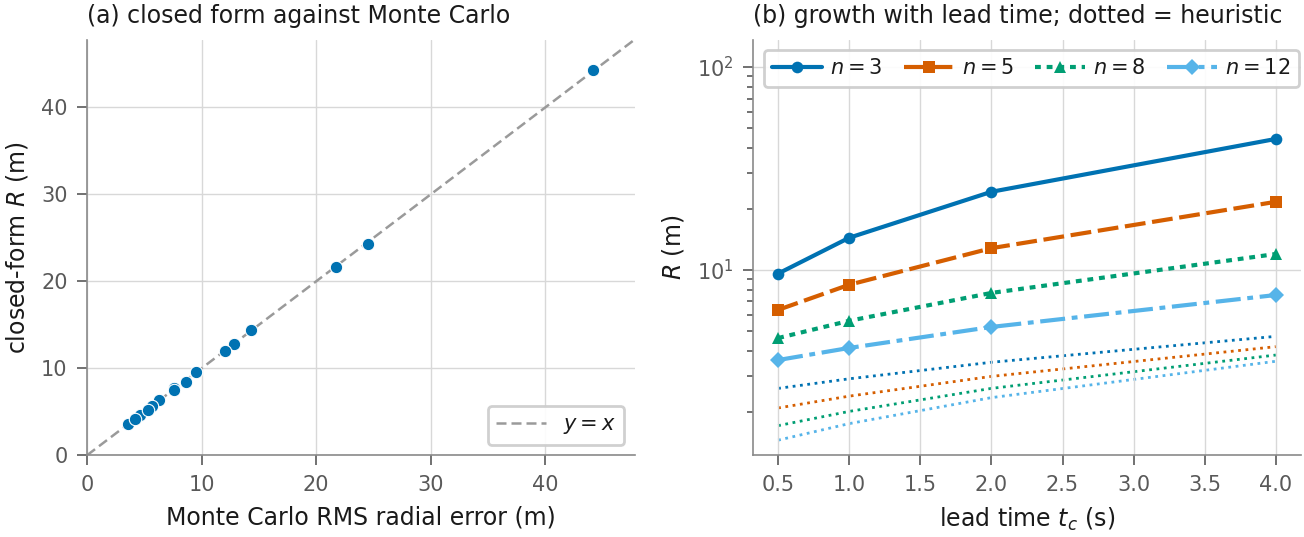}
\caption{(a) Closed-form $R$ against Monte Carlo RMS radial error across the
grid of Table~\ref{tab:e1}; the dashed line is $y=x$. (b) Growth of the
uncertainty with lead time $t_{c}$; dotted curves are the hand-tuned heuristic,
whose error grows with $t_{c}$.}
\label{fig:fig1}
\end{figure*}

For comparison the table carries a hand-tuned heuristic of the form
$\sigma_{s}/\sqrt{n}+0.15\,\sigma_{s}t_{c}$, which captures the direction of
every effect and none of their sizes. It understates the true uncertainty by
between $2.1\times$ and $9.4\times$, and the understatement grows with
$t_{c}$ --- which is the $b_{k}^{2}\mathrm{Var}(t_{c})$ term reasserting
itself. Fig.~\ref{fig:fig1} shows both comparisons.

\begin{table}[t]
\centering
\caption{Closed-Form Radial Uncertainty vs.\ Monte Carlo ($\sigma_{s}=4$ m,
$4\,000$ Trials per Cell)}
\label{tab:e1}
\begin{tabular}{rrrrrrr}
\toprule
$n$ & $t_{c}$ (s) & $R$ (m) & MC (m) & err.\ (\%) & heur.\ (m) & calib.\ (\%) \\
\midrule
 3 & 0.5 &  9.60 &  9.54 & 0.7 & 2.61 & 64.3 \\
 3 & 1.0 & 14.42 & 14.28 & 1.0 & 2.91 & 64.1 \\
 3 & 2.0 & 24.29 & 24.53 & 1.0 & 3.51 & 63.7 \\
 3 & 4.0 & 44.25 & 44.14 & 0.2 & 4.71 & 63.4 \\
 5 & 0.5 &  6.36 &  6.32 & 0.7 & 2.09 & 64.1 \\
 5 & 1.0 &  8.46 &  8.62 & 1.8 & 2.39 & 62.9 \\
 5 & 2.0 & 12.81 & 12.81 & 0.0 & 2.99 & 62.7 \\
 5 & 4.0 & 21.68 & 21.73 & 0.2 & 4.19 & 62.4 \\
 8 & 0.5 &  4.62 &  4.62 & 0.0 & 1.71 & 62.9 \\
 8 & 1.0 &  5.62 &  5.64 & 0.2 & 2.01 & 63.3 \\
 8 & 2.0 &  7.71 &  7.62 & 1.1 & 2.61 & 64.0 \\
 8 & 4.0 & 11.99 & 12.02 & 0.2 & 3.81 & 63.6 \\
12 & 0.5 &  3.60 &  3.59 & 0.2 & 1.45 & 63.4 \\
12 & 1.0 &  4.13 &  4.19 & 1.4 & 1.75 & 61.7 \\
12 & 2.0 &  5.24 &  5.33 & 1.6 & 2.35 & 62.1 \\
12 & 4.0 &  7.53 &  7.59 & 0.7 & 3.55 & 62.6 \\
\bottomrule
\end{tabular}
\end{table}

\section{Coverage Laws}\label{sec:coverage}

\subsection{Why the area ratio is not enough}

A natural way to size the mesh is an area ratio: to cover the disc of radius
$m\sigma$ with non-overlapping discs of radius $r$, take
$N\approx(m\sigma)^{2}/r^{2}$. This is optimistic, because it assumes the discs
tile the plane perfectly. They cannot: the densest packing of equal circles
covers only $\eta_{2}=\pi/\sqrt{12}$ of the plane~\cite{fejestoth1953}.

\subsection{The planar law}

Carrying the packing density through, $N$ touching discs of radius $r$ occupy a
roughly circular region of area $N\pi r^{2}/\eta_{2}$, that is, out to radius
$\rho=r\sqrt{N/\eta_{2}}$. Within that region a fraction $\eta_{2}$ of the area
is covered, and the enclosed mass of an isotropic Gaussian is
$1-\exp(-\rho^{2}/2\sigma^{2})$. Hence
\begin{equation}
p(N)=\eta_{2}\left(1-\exp\!\left(-\frac{Nr^{2}}{2\eta_{2}\sigma^{2}}\right)\right),
\label{eq:p2}
\end{equation}
which inverts to
\begin{equation}
N(p)=-\frac{2\eta_{2}\sigma^{2}}{r^{2}}\,\ln\!\left(1-\frac{p}{\eta_{2}}\right).
\label{eq:n2}
\end{equation}

Two consequences follow immediately. The count scales with the \emph{square} of
$\sigma/r$, with only the constant depending on the target coverage. The exponent is
exact by construction. Fitting it back from the closed form
would therefore be circular; inverting Monte Carlo measurements instead --- for
each $(\sigma,r)$, the smallest $N$ whose \emph{measured} coverage reaches a
target, over a grid of nine $(\sigma,r)$ pairs spanning $\sigma/r$ from
$1.0$ to $6.2$ --- gives a fitted exponent of $1.808$. The departure from $2$
sits entirely at the smallest budgets, where an integer number of discs cannot
track a continuous law: at $\sigma/r=1$ the measured requirement is $6$ agents
against the $3.9$ of \eqref{eq:n2}, while from $\sigma/r=3$ upward, where the
mesh is large enough for the continuum to apply, the two agree to within
$3\%$. And $p(N)\to\eta_{2}$ as $N\to\infty$, so no
arrangement of non-overlapping discs, however large, covers more than $90.69\%$
of the density. The area-ratio rule, which permits $p\to1$, lacks this ceiling
and is therefore not merely imprecise but qualitatively wrong in the regime
where it matters most.

\subsection{Validation and audit}

\begin{figure*}[t]
\centering
\includegraphics[width=0.92\textwidth]{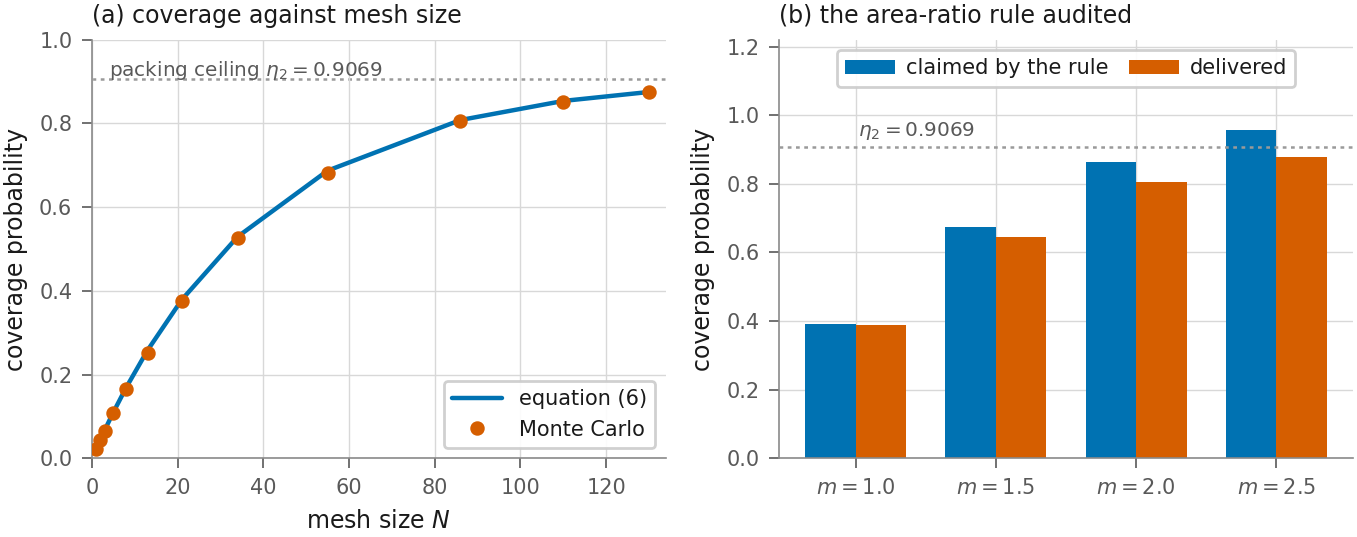}
\caption{(a) Coverage probability against mesh size: Monte Carlo against
\eqref{eq:p2}, with the packing ceiling $\eta_{2}$. (b) The area-ratio rule
audited --- claimed against delivered coverage at the $N$ it prescribes.}
\label{fig:fig2}
\end{figure*}

Against Monte Carlo using the production hexagonal-lattice generator,
\eqref{eq:p2} is accurate to $0.007$ absolute across $N\in[1,130]$
(Fig.~\ref{fig:fig2}a). Table~\ref{tab:e2} audits the area-ratio rule at
$\sigma/r=4.6$: what it claims, what it delivers, and what $N$ would actually
be required. At $m=2$ it claims $86.5\%$ and delivers $80.6\%$, and reaching
the claimed value needs $119$ agents rather than $86$, an inflation of
$1.38\times$. At $m=2.5$ it claims $95.6\%$, a figure above the packing ceiling
and therefore unattainable at any $N$.

\begin{table}[t]
\centering
\caption{The Area-Ratio Rule $N=(m\sigma)^{2}/r^{2}$, Audited
($\sigma=9.25$ m, $r=2$ m, $200\,000$ Trials per Cell)}
\label{tab:e2}
\begin{tabular}{rrrrr}
\toprule
$m$ & $N$ by rule & claimed $p$ & actual $p$ & $N$ required \\
\midrule
1.0 &  22 & 0.393 & 0.388 &  22.1 \\
1.5 &  49 & 0.675 & 0.646 &  53.0 \\
2.0 &  86 & 0.865 & 0.806 & 119.0 \\
2.5 & 134 & 0.956 & 0.877 & unreachable \\
\bottomrule
\end{tabular}
\end{table}

\subsection{The volumetric law}

\begin{figure}[t]
\centering
\includegraphics[width=0.85\columnwidth]{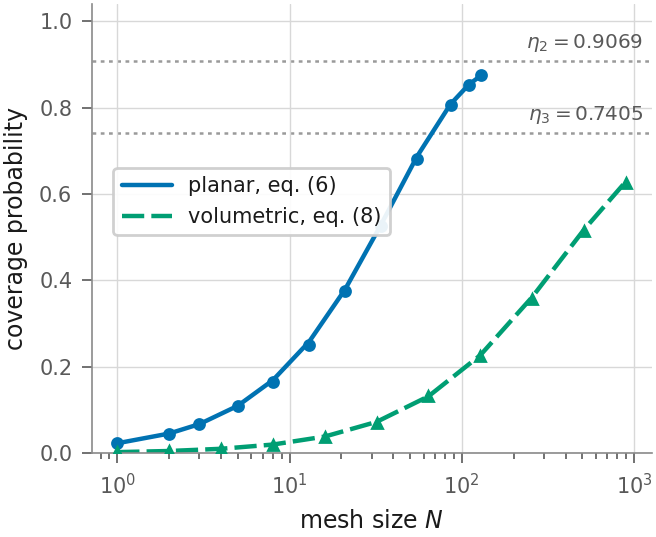}
\caption{Planar and volumetric coverage laws. Dotted lines mark the packing
ceilings $\eta_{2}=0.9069$ and $\eta_{3}=0.7405$.}
\label{fig:fig3}
\end{figure}

If the arrival altitude is also uncertain, the mesh must cover a volume. The
derivation carries over with two substitutions: $\eta_{3}=\pi/\sqrt{18}$ for
the densest sphere packing~\cite{hales2005,hales2017}, and the chi distribution
with three degrees of freedom for the enclosed mass,
\begin{equation}
F_{3}(\rho;\sigma)=\mathrm{erf}\!\left(\frac{\rho}{\sigma\sqrt{2}}\right)
 -\sqrt{\frac{2}{\pi}}\,\frac{\rho}{\sigma}\,
 \exp\!\left(-\frac{\rho^{2}}{2\sigma^{2}}\right),
\label{eq:chi3}
\end{equation}
with $\rho=r(N/\eta_{3})^{1/3}$ and $p_{3}(N)=\eta_{3}F_{3}(\rho;\sigma)$.
Against Monte Carlo over a face-centred-cubic lattice this is accurate to
$0.005$ absolute (Fig.~\ref{fig:fig3}). The exponent on $\sigma/r$ is
exactly $3$, again by construction rather than by measurement: $\rho$
scales as $(N/\eta_{3})^{1/3}$, so inverting at fixed $p$ gives a cubic
dependence identically. The empirical content here is the agreement of
$p_{3}(N)$ with Monte Carlo, not the exponent. The volumetric ceiling,
under the same separation assumption as before, is $74.05\%$.

The practical significance is the ratio. At $\sigma/r=4.6$ and $p=0.5$, a
planar mesh needs $31$ agents and a volumetric one $466$ --- a factor of $15$.
Constraining the arrival to a known plane is worth more than any plausible
improvement in the agents themselves.

\subsection{The ceiling belongs to the constraint, not to the problem}
\label{sec:regimes}

Everything above assumes $s\ge2r$. That assumption is doing more work than it
appears to, and it is worth isolating.

If the platform tolerates $s<2r$, coverage discs may overlap and the
arrangement is no longer a packing but a \emph{covering}. The governing
constant changes accordingly: the thinnest covering of the plane by equal
circles has density $\theta_{2}=2\pi/\sqrt{27}\approx1.2092$, established by
Kershner~\cite{kershner1939}. Because $\theta_{2}>1$, overlap wastes area ---
$N$ discs now blanket a region of area only $N\pi r^{2}/\theta_{2}$ --- but the
blanket has no gaps, so
\begin{equation}
p_{\mathrm{cov}}(N)=1-\exp\!\left(-\frac{Nr^{2}}{2\theta_{2}\sigma^{2}}\right),
\label{eq:pcov}
\end{equation}
with no leading constant. There is no ceiling: $p_{\mathrm{cov}}\to1$.

The two regimes therefore cross. Packing wins while the wasted overlap costs
more than the gaps do, and covering wins once the gaps matter more.
Equating \eqref{eq:p2} and \eqref{eq:pcov} puts the crossover at
\begin{equation}
p^{\star}=0.8028,
\label{eq:crossover}
\end{equation}
or $N\approx84$ at $\sigma/r=4.6$. Below $p^{\star}$ the packing law of
\eqref{eq:p2} is the right tool and $\eta_{2}$ is a genuine bound. Above it,
insisting on disjoint discs is simply a worse arrangement, and a design that
needs $p=0.95$ should abandon the separation constraint rather than conclude
the target is unattainable.

This qualifies the headline of Section~\ref{sec:coverage} rather than
overturning it. The area-ratio audit of Table~\ref{tab:e2} stands: within the
packing regime, that rule is optimistic and the $m=2.5$ target is genuinely
unreachable. But ``no mesh exceeds $90.69\%$'' is a statement about meshes
constrained to $s\ge2r$, and we state it that way. We note also that the
separation constraint in our own configuration is well below $2r$, so the
packing regime there reflects how the analytic policy chooses to arrange
itself, not a limit imposed by the vehicles.

\subsection{Which lever matters}

Because $N\propto\sigma^{2}/r^{2}$, halving $\sigma$ and doubling $r$ are
equally valuable, each reducing the required count fourfold, and no property of
the placement algorithm changes the exponent. Placement quality determines how
closely a team approaches this bound; it cannot move the bound.

Table~\ref{tab:e3} makes the point concretely on the sensing side, holding the
repositioning window fixed so that $t_{c}$ is constant and only the fix count
varies. Raising the cadence from $3$ to $24$ fixes tightens $\sigma$ by
$2.1\times$ and reduces the mesh required for $80\%$ coverage from $29$ agents
to $7$. The competing lever behaves as the exponent dictates: at fixed cadence,
moving $r$ from $1$ m to $8$ m takes the requirement from $184$ agents to $3$.

\begin{table}[t]
\centering
\caption{Mesh Cost vs.\ Sensor Cadence ($\sigma_{s}=4$ m, $r=3$ m,
Repositioning Window Held at $2.0$ s so That $t_{c}$ Is Constant)}
\label{tab:e3}
\begin{tabular}{rrrrrr}
\toprule
fixes & rate (Hz) & $R$ (m) & MC (m) & $\sigma$ (m) & $N$ at $p=0.8$ \\
\midrule
 3 &  0.90 & 11.68 & 11.53 & 8.26 & 29.4 \\
 4 &  1.35 & 11.00 & 10.80 & 7.78 & 26.1 \\
 5 &  1.79 & 10.34 & 10.23 & 7.31 & 23.0 \\
 6 &  2.24 &  9.75 &  9.84 & 6.90 & 20.5 \\
 8 &  3.14 &  8.79 &  8.86 & 6.22 & 16.7 \\
12 &  4.93 &  7.47 &  7.62 & 5.28 & 12.0 \\
16 &  6.73 &  6.60 &  6.77 & 4.67 &  9.4 \\
24 & 10.31 &  5.50 &  5.50 & 3.89 &  6.5 \\
\bottomrule
\end{tabular}
\end{table}

\section{Is Hexagonal Packing the Right Arrangement?}\label{sec:strategies}

Equation~\eqref{eq:p2} assumes a hexagonal arrangement, justified by Thue's
theorem. That theorem concerns covering the infinite plane, whereas our mesh
covers a finite Gaussian with a finite budget. The assumption therefore
deserves an empirical check, and the check is informative about the relation
between this problem and classical coverage control.

We compare five arrangements at matched $N$, $\sigma$ and $r$: the hexagonal
lattice used throughout; a Fermat-spiral (phyllotaxis) arrangement, which is
locally near-hexagonal but free of grid artefacts; a square lattice at the same
spacing, with packing density $\pi/4$; a centroidal Voronoi configuration
obtained by Lloyd's algorithm on samples from the density, which is what
coverage control converges to~\cite{cortes2004,du1999}; and uniform random
placement subject to non-overlap, as a weak reference.

\begin{table}[t]
\centering
\caption{Coverage by Placement Strategy at Matched Budget
($\sigma=9.25$ m, $r=2$ m, $200\,000$ Trials per Cell). Random Placement
Draws Uniformly in a Disc, Rejecting Any Agent Within $2r$ of Another, the
Disc Grown Until All $N$ Fit}
\label{tab:e7}
\begin{tabular}{lrrrr}
\toprule
strategy & $N{=}9$ & $N{=}25$ & $N{=}60$ & $N{=}130$ \\
\midrule
hexagonal lattice        & \textbf{0.185} & \textbf{0.422} & \textbf{0.712} & \textbf{0.874} \\
Fermat spiral            & 0.178 & 0.404 & 0.669 & 0.821 \\
square lattice           & 0.185 & 0.408 & 0.651 & 0.768 \\
Lloyd / CVT              & 0.094 & 0.219 & 0.508 & 0.852 \\
random, non-overlapping  & 0.175 & 0.339 & 0.480 & 0.524 \\
\bottomrule
\end{tabular}
\end{table}

\begin{figure*}[t]
\centering
\includegraphics[width=0.92\textwidth]{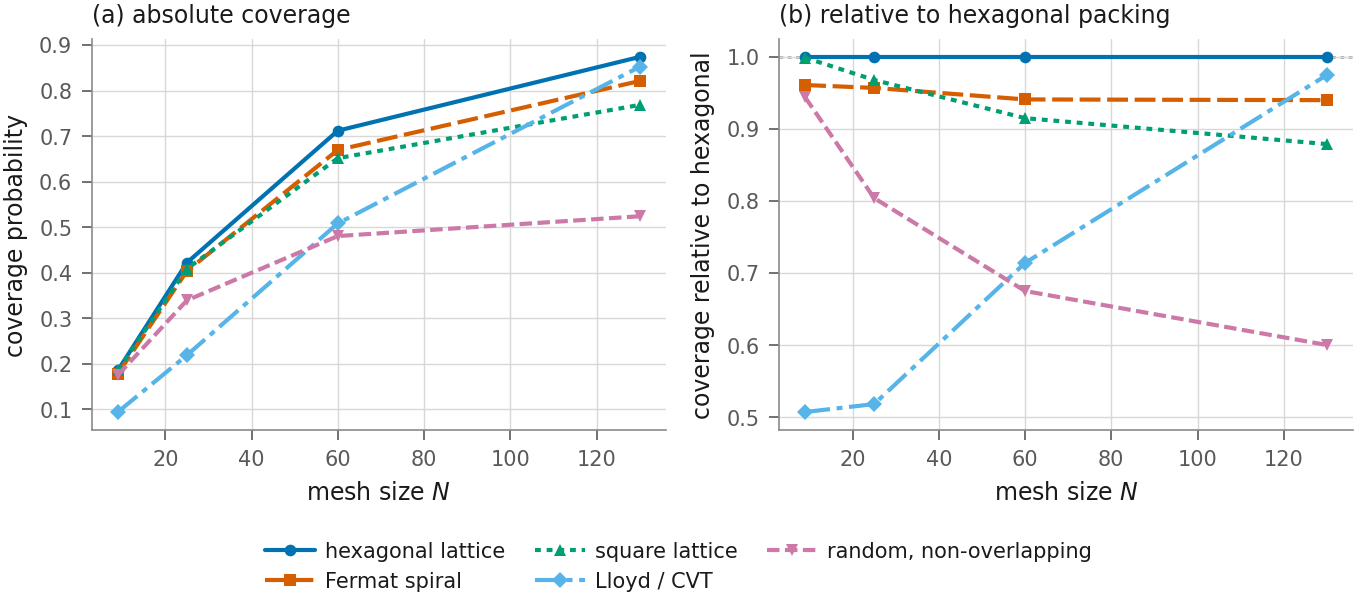}
\caption{Placement strategies at matched budget. (a) Absolute coverage.
(b) Coverage relative to hexagonal packing. The centroidal Voronoi
configuration is worst where the budget is tightest.}
\label{fig:fig5}
\end{figure*}

Table~\ref{tab:e7} and Fig.~\ref{fig:fig5} confirm the assumption: hexagonal
packing is best at every budget tested, tied only by the square lattice at
$N=9$ where the arrangement is too small for the difference in density to
express itself. The Fermat spiral trails it by a steady $4$ to $6\%$, and the
square lattice degrades as $N$ grows, consistent with its lower packing
density.

The centroidal Voronoi configuration is the interesting case. At $N=9$ it
attains $0.507$ of hexagonal coverage and at $N=25$ only $0.518$, converging to
$0.975$ by $N=130$. The reason is not an implementation artefact but the
objective: Lloyd's algorithm minimises mean squared distance to the nearest
agent, which rewards placing agents out in the tails of the density where the
squared-distance penalty is largest. Binary containment gives no credit for
being closer to a point one does not cover. Measured nearest-neighbour spacing
confirms it --- at $N=60$ the CVT solution rests at $5.56$ m against the $4.0$ m of a
touching pack, spending budget on reach it cannot convert into coverage.

The practical reading is that the standard tool for this family of problems is
the wrong tool for this member of it, and most wrong exactly where budgets are
tight. It also sharpens the claim of Section~\ref{sec:learning}: the analytic
baseline there is not merely a convenient heuristic but the best of five
arrangements including the field's default.

\section{Robustness of the Error Model}\label{sec:robustness}

Equation~\eqref{eq:p2} assumes an isotropic Gaussian error. Two departures are
likely in practice. Heavy tails arise from occasional bad fixes, missed
manoeuvres or track breaks. Anisotropy arises because crossing geometry rarely
leaves the two in-plane axes equally determined. A third, bias, arises when the
predictor is systematically off-centre rather than merely imprecise --- which
is what an unmodelled manoeuvre produces.

We size a mesh \emph{as if} the error were Gaussian, which is what a
practitioner holding only $\sigma$ would do, then measure what that mesh
delivers against each true distribution. The normalisation of the shape
departures matters and is stated explicitly: the Student-$t$ cases hold the
marginal variance at $\sigma^{2}$, and the elliptical cases hold the trace
of the covariance at $2\sigma^{2}$, so that the RMS radial error matches the
isotropic case and only the shape differs. The marginals cannot themselves
be held fixed under anisotropy, since one axis must widen as the other
narrows; under a determinant-preserving normalisation instead, the minor
axis would contract less and the sign of the anisotropy result would change.

\begin{table}[t]
\centering
\caption{Coverage Delivered by a Mesh Sized for a Gaussian ($N=43$, Target
$0.60$, $\sigma=9.25$ m, $r=2$ m)}
\label{tab:e8}
\begin{tabular}{lrrr}
\toprule
error model & delivered & surplus & $N$ for target \\
\midrule
Gaussian (baseline)  & 0.601 & $+$0.001 &  43 \\
Student-$t$, $\nu=10$ & 0.625 & $+$0.025 &  39 \\
Student-$t$, $\nu=5$  & 0.662 & $+$0.062 &  35 \\
Student-$t$, $\nu=3$  & 0.734 & $+$0.134 &  25 \\
elliptical 2:1       & 0.625 & $+$0.025 &  40 \\
elliptical 4:1       & 0.626 & $+$0.026 &  37 \\
elliptical 8:1       & 0.634 & $+$0.034 &  37 \\
biased $0.5\sigma$   & 0.562 & $-$0.038 &  47 \\
biased $1.0\sigma$   & 0.457 & $-$0.143 &  67 \\
biased $2.0\sigma$   & 0.187 & $-$0.413 & 138 \\
\bottomrule
\end{tabular}
\end{table}

\begin{figure*}[t]
\centering
\includegraphics[width=0.92\textwidth]{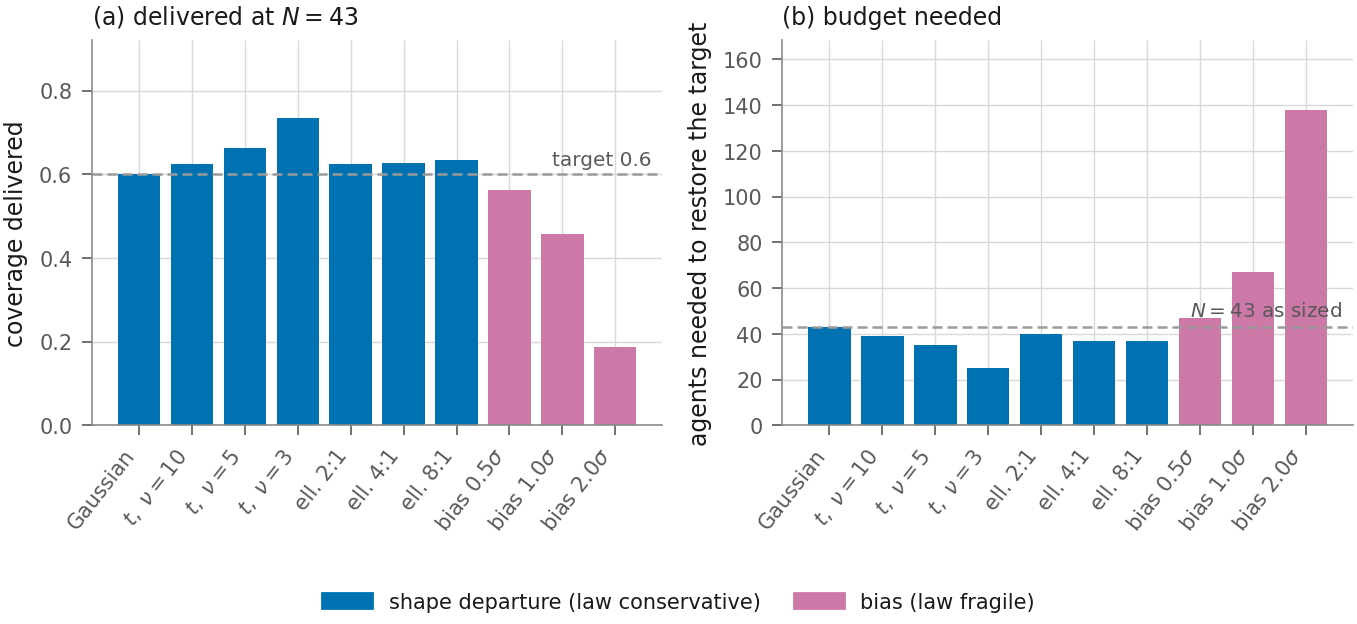}
\caption{Robustness to a misspecified error model. (a) Coverage delivered at a
budget sized for a Gaussian. (b) Budget needed to restore the target. Shape
departures (blue) leave the law conservative; bias (pink) does not.}
\label{fig:fig6}
\end{figure*}

The result splits cleanly, and not in the direction we expected.

\textbf{Shape departures make the law conservative.} At fixed marginal
$\sigma$, both heavy tails and anisotropy concentrate more mass near the mode
than a Gaussian does, so a mesh sized by \eqref{eq:n2} over-delivers. A
Student-$t$ with $\nu=3$ reaches the target with $25$ agents where the Gaussian
law prescribes $43$. This is a statement about shape at fixed spread, and it
does not license ignoring the error model: if the true $\sigma$ exceeds the
fitted one, the quadratic law applies to the true value and the cost is
quadratic in the discrepancy.

\textbf{Bias is the departure that breaks it.} A one-sigma offset of the aim
point costs $14$ points of coverage, a two-sigma offset $41$. Restoring the
target takes $67$ and $138$ agents respectively, against $43$ --- a
$3.2\times$ inflation purchased purely to compensate for a mesh centred in the
wrong place. This matters because Section~\ref{sec:commit} shows that
unmodelled manoeuvre displaces the crossing point rather than blurring it, so
it enters the budget here rather than as extra spread. The engineering reading
is unambiguous: budget spent on the aim point dominates budget spent on team
size, by a wide margin.

\section{The Commit-Time Window}\label{sec:commit}

\subsection{The trade}

Sections~\ref{sec:prediction} and~\ref{sec:coverage} price a mesh given
$\sigma$. A team that must physically reposition faces a further constraint,
because $\sigma$ is not exogenous: it depends on how early the team commits.

Let $T$ be the commit time --- the moment the team stops updating its aim point
and flies to the formation. $T$ is squeezed from both sides. Commit late and
the prediction is sharp but there is no time to fly. Commit early and there is
ample transit time but the arrival point must be extrapolated further.

Two contributions enter $\sigma(T)$ in quadrature. The sensor term is
\eqref{eq:R}, which \emph{falls} as the track lengthens. The second is
unmodelled lateral acceleration: the fit assumes a straight line, and an
acceleration $\alpha$ displaces the crossing point by $\alpha T^{2}/2$, growing
quadratically in the lead. We parameterise it by a per-axis standard deviation
in units of $g$; this is a crude stand-in for the correlated-acceleration
models of the tracking literature~\cite{singer1970,li2003}, adequate because
the conclusion turns on which term dominates rather than on its exact constant.

Including this second term is essential and its omission is not benign. In the
configuration below --- a track of roughly $30$ s at $2.5$ Hz --- the sensor
term alone gives $\sigma\approx1$ m, which would imply a single agent suffices.
Fig.~\ref{fig:fig4}a shows the actual budget: the sensor term stays between
$0.96$ and $1.59$ m across the whole range of $T$, while the manoeuvre term
runs from $2.5$ m to $481$ m. Two orders of magnitude separate them. The carefully
derived sensor model of Section~\ref{sec:prediction} is, at these lead times,
not the binding term.

\subsection{An empty window}

Feasibility requires that the worst assigned agent reach its slot. With a
formation half-extent $L=r\sqrt{N/\eta_{2}}$ and a corridor of half-width $C$
within which the formation centre may fall, the worst assigned transit is
$\sqrt{2}(L+C)$, while an agent that must start and end at rest covers
\begin{equation}
d(T)=\begin{cases}
 aT^{2}/4, & T\le 2v_{\max}/a,\\[2pt]
 v_{\max}T-v_{\max}^{2}/a, & \text{otherwise.}
\end{cases}
\label{eq:reach}
\end{equation}
Note the two ramps: an agent that must arrive \emph{and hold station} brakes as
well as accelerates, so the familiar $v_{\max}T-v_{\max}^{2}/2a$ overstates
reach by one ramp's worth.

\begin{figure*}[t]
\centering
\includegraphics[width=0.92\textwidth]{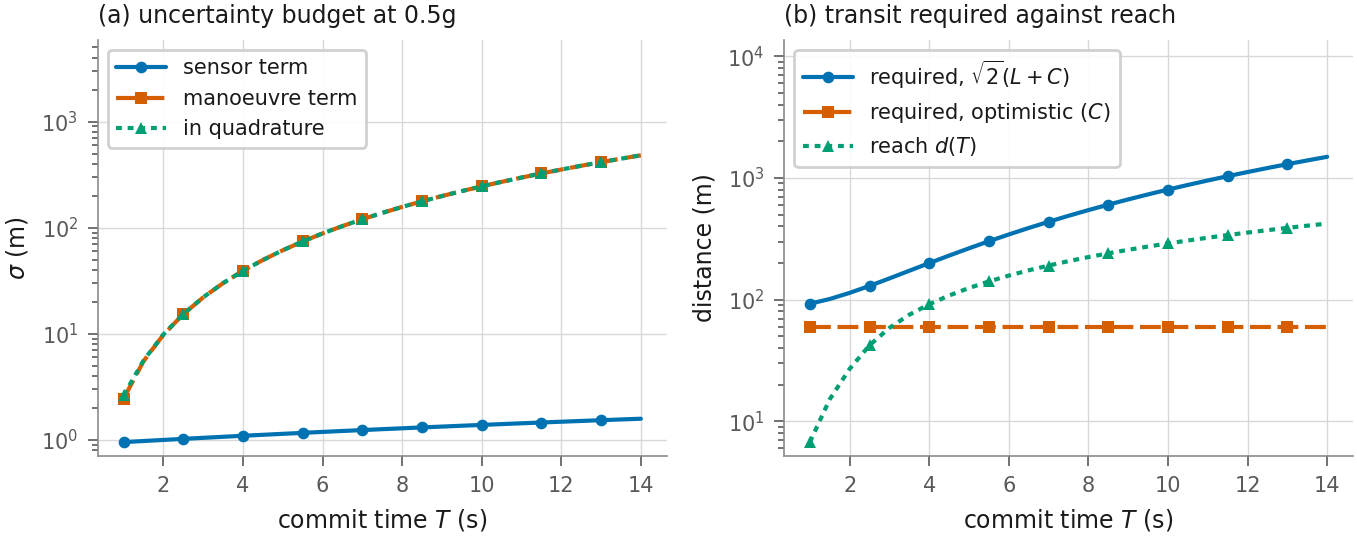}
\caption{(a) The uncertainty budget against commit time: the sensor term is
negligible beside unmodelled manoeuvre at every $T$. (b) Transit required
against reach available; the two bounds on required transit bracket the answer,
and neither admits a useful window.}
\label{fig:fig4}
\end{figure*}

For $C=60$ m, $r=6$ m, $v_{\max}=33$ m/s, $a=27$ m/s$^{2}$ and a manoeuvre
uncertainty of $0.5g$, \textbf{no $T$ satisfies both constraints}. The closest
approach is at $T=1.0$ s, short by $86$ m of transit. The reason is visible in
Fig.~\ref{fig:fig4}b: at small $T$ the required transit is dominated by the
corridor itself, $\sqrt{2}C\approx85$ m, which the agent cannot cross in the
time available regardless of how small the formation is.

Relaxing the transit model to an optimistic bound --- assuming agents are
already distributed around the formation centre, so that typical transit is $C$
rather than $\sqrt{2}(L+C)$ --- opens a window at $T\geq3.5$ s. But the
cheapest feasible mesh in that window costs $97$ agents. The static
alternative fares no better. Covering a fraction $p$ of a corridor of
radius $C$ with disjoint discs needs $N=pC^{2}/r^{2}$, here $80$ agents,
but the corridor cannot hold them: walking the lattice outward until a
disc would cross the boundary admits only $73$, capping static coverage
at $0.730$. At the stated target of $0.80$ this configuration therefore
admits no solution at all --- neither repositioning nor blanketing.

Table~\ref{tab:e5} sweeps corridor width against manoeuvre uncertainty. A
window exists only in the left-hand column, at $0.1g$ --- that is, only if the
tracked point flies very nearly straight. At $0.5g$ and above the window is
empty at every corridor width tested.

Stated in the terms of the application, this is the sharpest engineering result
in the paper. A mesh that repositions onto a prediction can defend a site
against an object that flies nearly straight, and cannot defend it against one
that manoeuvres: the boundary sits between $0.1g$ and $0.5g$ at every corridor
width we tested. The limit belongs to the geometry rather than to the vehicles.
At small commit times the transit an agent must fly is dominated by the
corridor itself, $\sqrt{2}C\approx85$ m, which is already beyond reach before
the formation's own extent is counted; closing the $86$ m shortfall at the best
commit time found would take an acceleration near $18g$. Faster or more agile
interceptors therefore do not open the window at any plausible margin. A
narrower corridor, an earlier and better prediction, or a larger engagement
radius do --- which is where the design effort belongs.

\begin{table}[t]
\centering
\caption{Cheapest Feasible Mesh Size; ``---'' Where the Commit Window Is
Empty. Counts Are Continuous, per Section~\ref{sec:problem}. Commit Times
Are Evaluated on a $0.5$ s Grid; the Cheapest Mesh Sits at the Smallest
Feasible $T$, So the Step Matters}
\label{tab:e5}
\begin{tabular}{rrrrr}
\toprule
corridor $C$ & $0.1g$ & $0.5g$ & $1.0g$ & $3.0g$ \\
\midrule
$\pm15$ m &   1.1 & --- & --- & --- \\
$\pm25$ m &   2.2 & --- & --- & --- \\
$\pm40$ m &   4.0 & --- & --- & --- \\
$\pm60$ m &  16.3 & --- & --- & --- \\
$\pm90$ m & 106.3 & --- & --- & --- \\
\bottomrule
\end{tabular}
\end{table}

\section{Applying the Bounds}\label{sec:recipe}

The results above are individually simple, and their value is in the order they
are applied; we set out that order below.

\textbf{Step 1: price the prediction, including what the fit does not model.}
Compute $R$ from \eqref{eq:R} for the intended cadence and lead time, then add
the manoeuvre term of Section~\ref{sec:commit} in quadrature. Section~\ref{sec:commit}
is emphatic about the order of magnitude involved: a long track drives the
sensor term to a metre or two while the manoeuvre term reaches hundreds, so a
budget containing only the first is not conservative but simply wrong. If the
tracked point may accelerate at all, the manoeuvre term is the budget.

\textbf{Step 2: convert the budget into a team size.} Apply \eqref{eq:n2}, not
an area ratio. Check the target against the ceiling first: while $s\ge2r$, any
$p\ge\eta_{2}$ is unattainable at every $N$ (if the platform tolerates
$s<2r$, use \eqref{eq:pcov} instead and there is no ceiling), and an area ratio will nonetheless return a
finite number. If the arrival altitude is also free, use the volumetric form
and expect roughly an order of magnitude more agents.

\textbf{Step 3: check that the team can physically be where the law puts it.}
The count from Step 2 implies a formation half-extent $L=r\sqrt{N/\eta_{2}}$,
which with the corridor half-width sets a transit distance. Compare it against
\eqref{eq:reach} at the intended commit time. If no commit time satisfies both
constraints, no placement policy can succeed and the configuration must change
--- which is a conclusion available in seconds, before any controller exists.

\textbf{Step 4: compare against blanketing.} Repositioning is not automatically
worthwhile. A static mesh covering a fraction $p$ of the corridor costs $pC^{2}/r^{2}$
agents and needs no prediction, no commit decision and no transit. Check it
against capacity as well as against cost: the corridor holds at most
$\eta_{2}C^{2}/r^{2}$ discs asymptotically and fewer once the boundary is
accounted for, so a target above that capacity is unreachable statically
however many agents are available. In Section~\ref{sec:commit}'s configuration it is cheaper in nominal count
($80$ against $97$) yet fails the capacity check, which is exactly why the
check belongs in this step: neither option is feasible at the stated target.
The comparison takes two lines and settles whether the repositioning concept is
worth building at all.

\textbf{Step 5: decide where to spend.} Because $N\propto\sigma^{2}/r^{2}$,
halving $\sigma$ and doubling $r$ are worth a factor of four each, and no
placement algorithm changes that exponent. Section~\ref{sec:robustness} adds a
sharper version: bias in the aim point costs more than spread. A one-sigma
offset demands $1.6\times$ the agents and a two-sigma offset $3.2\times$, so
effort spent on removing a systematic aiming error dominates effort spent on
fleet size, and dominates it by more than the quadratic law alone suggests.

\textbf{Step 6: only now consider learning.} Steps 1--5 produce a feasibility
ceiling, a null-policy score and a classical baseline. These are the three
numbers that make a learned result interpretable, and
Section~\ref{sec:strategies} supplies a fourth consideration: the classical
baseline should be a packed arrangement, since the centroidal Voronoi
configuration that coverage control would naturally supply is the wrong
objective for binary containment and loses half its coverage at tight budgets.

Applied in this order, the analysis of a new configuration costs minutes. The
sequence also makes clear which failures are worth investigating: a policy
scoring near the ceiling is not underperforming, and a policy scoring near the
null score on a task whose ceiling is high is a genuine finding about the
learner rather than about the task.

\section{Implications for Learned Placement}\label{sec:learning}

\subsection{Measured outcomes}

The bounds above were derived after, not before, a sequence of
reinforcement-learning experiments on this task in an accompanying prototype.
We report those outcomes because the sequence itself is the argument.

The geometry is the corrected one of Section~\ref{sec:faults}: $25$ agents,
sector $\pm35$ m, corridor $\pm20$ m, $r=3$ m, mean $\sigma=4.57$ m, a $4$ s
window. All policies are evaluated through the identical environment code path
over the same $100$ episode seeds, with percentile bootstrap
intervals~\cite{efron1993} over episodes.

Before the comparison, the denominator --- and here the analysis of
Section~\ref{sec:regimes} has to be applied to our own configuration rather
than left as a general remark.

A crossing falls inside the reachable sector in every episode, so reachability
contributes a factor of $1.000$. For the coverage factor, the environment's
separation limit is $s=2.5$ m against $2r=6$ m. By Section~\ref{sec:regimes}
the packing law therefore does \emph{not} apply: discs may overlap, and the
governing constant is the covering density. The two laws give $0.904$
by \eqref{eq:p2} and $0.988$ by \eqref{eq:pcov} for $25$ agents at
$\sigma=4.57$ m and $r=3$ m, so the applicable ceiling is
$1.000\times0.988=0.988$. Table~\ref{tab:e6} reports both columns.

Two consequences follow. Quoting the packing ceiling would have flattered
every policy by roughly eight percentage points. And the packing ceiling of
$0.904$ sits \emph{above} the crossover $p^{\star}=0.8028$ of
\eqref{eq:crossover}, so by this paper's own analysis the hexagonal
arrangement is not optimal at this operating point: it is the best of the five
\emph{packed} arrangements of Section~\ref{sec:strategies}, and it is what the
analytic policy implements, but an overlapping arrangement would do better. We
did not notice this until the covering regime was worked out.

\begin{table}[t]
\centering
\caption{Placement Policies Over $100$ Episode Seeds Through an Identical
Environment Path. Intervals Are $95\%$ Percentile Bootstrap Over Episodes;
PPO Is Reported at Three Trainer Seeds. The Covering Column Is the Applicable
One, Since $s<2r$}
\label{tab:e6}
\begin{tabular}{lccrr}
\toprule
 & & & \multicolumn{2}{c}{of ceiling} \\
\cmidrule(l){4-5}
policy & hit rate & coll. & packing & covering \\
\midrule
null (static lattice)       & 0.14 [0.08, 0.21] & \textbf{0.00} & 15.5\% & 14.2\% \\
\midrule
centralised PPO, seed 0     & 0.12 [0.06, 0.19] & 0.78 & 13.3\% & 12.1\% \\
centralised PPO, seed 1     & 0.19 [0.12, 0.27] & 0.74 & 21.0\% & 19.2\% \\
centralised PPO, seed 2     & 0.14 [0.08, 0.21] & 0.61 & 15.5\% & 14.2\% \\
\quad mean over seeds       & 0.15              & 0.71 & 16.6\% & 15.2\% \\
\midrule
analytic hex packing        & \textbf{0.84} [0.77, 0.91] & 0.62 & 92.9\% & \textbf{85.0\%} \\
\quad + transit repulsion   & 0.79 [0.71, 0.87] & 0.49 & 87.3\% & 79.9\% \\
\bottomrule
\end{tabular}
\end{table}

The learner was PPO~\cite{schulman2017} as implemented in
Stable-Baselines3~\cite{raffin2021}, with a two-layer $128\times128$ MLP
policy, learning rate $3\times10^{-4}$, $2048$ steps per rollout, minibatch
$256$, $\gamma=0.995$, trained for $4\times10^{5}$ environment steps at each
seed. No hyperparameter search was run.

Because every policy is evaluated on the same episodes, the episode is a
matched unit and the comparison against the null policy should be
\emph{paired}. Table~\ref{tab:paired} reports the bootstrap of the per-episode
difference. At all three trainer seeds the interval contains zero; for both
analytic policies it does not, by a wide margin. This is the statement the
paper relies on. Pairing is the comparison a matched design licenses, not
automatically a tighter one. Here the episode-level correlation between each
policy and the null policy runs from $-0.12$ to $+0.12$, so the paired
intervals are in fact slightly \emph{wider} than the marginals of
Table~\ref{tab:e6}. Pairing buys correctness rather than precision in this
instance; arguing from overlapping independent intervals would not have been
valid.

\begin{table}[t]
\centering
\caption{Paired Bootstrap of the Per-Episode Difference Against the Null
Policy, Same $100$ Episodes}
\label{tab:paired}
\begin{tabular}{lcc}
\toprule
policy & mean difference & $95\%$ CI \\
\midrule
centralised PPO, seed 0   & $-$0.020 & [$-$0.110, $+$0.070] \\
centralised PPO, seed 1   & $+$0.050 & [$-$0.060, $+$0.160] \\
centralised PPO, seed 2   & $+$0.000 & [$-$0.100, $+$0.100] \\
\midrule
analytic hex packing      & $\mathbf{+0.700}$ & [$+$0.600, $+$0.790] \\
\quad + transit repulsion & $+$0.650 & [$+$0.550, $+$0.750] \\
\bottomrule
\end{tabular}
\end{table}

Averaged over trainer seeds the learned policy scores $0.15$ against the null
policy's $0.14$, with a seed-to-seed spread from $0.12$ to $0.19$ that
straddles the null policy in both directions. On a single seed the conclusion
would have been either ``PPO is worse than doing nothing'' or ``PPO beats doing
nothing by a third'', depending which seed was drawn; this is the failure mode
\cite{colas2018} quantifies.

The analytic solution reaches $85.0\%$ of the applicable ceiling. By
Section~\ref{sec:strategies}, hexagonal packing is the best of the five
\emph{packed} arrangements at this budget, so the residual $15\%$ is not all
headroom for a smarter policy of the same kind: part is distance to the bound,
and part is the packing--covering gap identified above, which no packed
arrangement can close. PPO's explained variance never left the
neighbourhood of zero, ending at $-0.12$: the value function did not fit the
task at all.

PPO's collision rate of $61$--$78\%$ is high, but so is the analytic policy's
$62\%$, and the repulsion variant only brings it to $49\%$. Collisions are
therefore \emph{not} a discriminator between learning and the analytic method
here; both incur them and only the null policy, which never moves, avoids them
entirely. They arise in transit, when agents converging on a formation from
opposite sides of the sector cross paths, rather than in the settled formation
--- which is why a short-range repulsion term reduces them at a small cost in
coverage. A deployment would need to address this for either method, and we do
not claim the analytic solution is collision-safe.

An initial ``retrain'' at the released configuration reproduced the original
run to eight decimal places in every training statistic, because the trainer
seed is pinned in the configuration file. It is a determinism check and
nothing more; it is reported separately in the released records and excluded
from the mean above, since counting it would fabricate a replicate.

\subsection{Why this is unsurprising}

The placement problem is not an unstructured control problem. It decomposes
into two classical sub-problems with known solutions: where the formation
should sit, which is densest packing over a radially symmetric density; and
which agent occupies which slot, which is linear assignment~\cite{kuhn1955}. A
monolithic policy is asked to rediscover both in a $50$-dimensional continuous
action space whose parameterisation treats interchangeable agents as unrelated
output dimensions --- precisely the symmetry that makes the assignment
sub-problem tractable, and precisely what permutation-equivariant
architectures~\cite{battaglia2018,liu2020,vanderpol2022} exist to supply.

Concentration is overwhelmingly the right strategy: $84\%$ against $14\%$ for
holding station. What fails is learning it end to end.

\subsection{Three faults that mimic learning failure}\label{sec:faults}

Before the comparison above was meaningful, three specification faults had to
be removed. Each produced a symptom easily misread as ``the policy did not
learn''. They are instances of a general phenomenon~\cite{ng1999,skalse2022},
but the specific forms are worth recording.

\textbf{An infeasible scenario.} The first configuration had a ceiling near
$10\%$, decomposed as $56.2\%$ for the arrival falling inside the reachable
area times $18.1\%$ for the best-case coverage of nine agents at $r=2$ m and
$\sigma=9.25$ m. Five rounds of reward shaping were spent on a policy already
operating at or above what the configuration permitted; measured hit rates of
$4$--$16\%$ were between $40\%$ and $160\%$ of the ceiling. The correct first
diagnostic costs seconds and was performed only after the fact.

\textbf{Penalty terms of the wrong order in $N$.} Pairwise separation
violations were charged per unordered pair, which grows as $O(N^{2})$, while
the coverage reward was a mean over agents, which is $O(1)$. At $N=25$ the
penalty exceeded the entire achievable coverage reward by more than an order of
magnitude, making dispersal reward-optimal. The diagnostic signature was
explained variance pinned near $10^{-7}$ alongside a large value loss. Every
term in a multi-agent reward must carry the same order in $N$.

\textbf{A baseline that already solves the task.} After the previous fix, the
mesh was dense enough that the resting lattice scored $63\%$ with no control
input at all, making any learned result unfalsifiable. The corrected geometry
holds a wide sector sparsely, at a lattice spacing far above $r$, so that the
null policy scores $14\%$.

\subsection{The recommendation}

For constrained multi-agent coverage tasks we suggest three things be computed
and reported before any learning experiment, and quoted alongside every result:

\begin{enumerate}
\item the feasibility ceiling, as the product of reachability and best-case
      coverage;
\item the null-policy score, so that a task the null policy already solves is
      caught before training rather than after;
\item a closed-form or classical-algorithm baseline, evaluated through the same
      code path as the learner.
\end{enumerate}

A bare success rate without these is uninterpretable. In this project, three of
the four things that looked like learning failures were scenario or reward
specification faults, and the fourth --- the genuine one --- is explained by
Section~\ref{sec:commit} without reference to the learner at all.

\section{Limitations}\label{sec:limits}

\textbf{The manoeuvre model is crude.} Unmodelled acceleration enters as an
independent per-axis Gaussian with fixed standard deviation, integrated over
the lead time. A real manoeuvring point is correlated in time and bounded in
magnitude~\cite{singer1970,li2003}, and a constant-acceleration fit would
absorb part of what we charge to uncertainty. The qualitative conclusion ---
that this term dominates beyond a few seconds of lead --- is robust to the
details; the specific numbers in Table~\ref{tab:e5} are not.

\textbf{The transit bounds bracket rather than resolve.} We report a
worst-assignment bound and an optimistic bound because the true requirement
depends on the resting layout, which we do not model. The conclusion that
repositioning loses holds under both, but a layout-aware analysis would give a
single number.

\textbf{Point-mass agents.} Attitude dynamics, actuator lag and downwash
interactions are absent. In a separate full-physics run of the accompanying
prototype the formation settled $10$--$27$ m from its slots against the
$10$--$15$ m predicted here, so the planar model is optimistic but not by an
order of magnitude.

\textbf{Centralised control.} One policy emits all $N$ commands, and assignment
is computed globally. Decentralised operation on local information is the
natural next step; the assignment sub-problem admits distributed
solutions~\cite{bertsekas1988,zavlanos2008,choi2009}, but the packing
sub-problem would need a consensus mechanism we have not designed.

\textbf{The negative learning result concerns one architecture.} An
unstructured centralised policy loses to analytic packing across three seeds. A
permutation-equivariant policy with assignment supplied by the Hungarian
algorithm and only residual corrections learned has not been tried, and until
it has, the result is about end-to-end learning rather than about learned
placement in general. Given how strongly Section~\ref{sec:strategies} favours
the packed arrangement, we would expect a structured learner to approach it
rather than beat it.

\textbf{Single prediction.} The analysis computes one prediction and commits.
Continuous re-prediction with a shrinking $\sigma$ is the obvious extension and
would change the commit-time analysis qualitatively, since the team could
re-converge as the estimate sharpens.

\section{Conclusion}\label{sec:conclusion}

We have given closed forms, each validated numerically, for the quantities that
bound multi-agent coverage of an uncertain arrival point before any policy is
chosen: the propagated uncertainty of the prediction, accurate to $1.8\%$
against Monte Carlo and calibrated to within a point of nominal; the coverage
probability of a packed mesh in the plane and in a volume, accurate to $0.007$
and $0.005$ absolute, whose quadratic and cubic exponents are exact by
construction and whose counts agree with Monte Carlo inversion to within
$3\%$ once the mesh is large enough for the continuum to apply; and the
admissible window of commit times, which can be empty.

Four consequences follow. There is a ceiling on coverage ---
$90.69\%$ in the plane, $74.05\%$ in a volume --- that no arrangement of
mutually separated agents can exceed at any size, and an area-ratio rule that
misses it prescribes finite team sizes for unattainable targets. That ceiling
belongs to the separation constraint rather than to the problem: permit overlap
and the governing constant becomes the thinnest-covering
density~\cite{kershner1939}, above a crossover at $p=0.80$ this is the better
arrangement, and no ceiling remains --- so a design needing high coverage
should relax the constraint rather than declare the target impossible. The
arrangement also matters within the packing regime, and the standard tool is
the wrong one: a centroidal Voronoi configuration attains half the coverage of
a packed one at tight budgets, because it optimises distance rather than
containment. And the Gaussian assumption is conservative under heavy tails and
anisotropy but fragile under bias, which is exactly the form unmodelled
manoeuvre takes.

In the terms of the application that motivated the work, the answer is negative
and specific. A repositioning mesh of interceptors can defend a site against a
nearly ballistic object and cannot defend it against one manoeuvring at half a
$g$ or more; the binding constraint is the transit imposed by the corridor
width, not the agility of the vehicles, and no placement policy, learned or
analytic, can lift it.

The methodological claim is the one we would most like to see adopted. A
success rate reported for a hand-designed multi-agent task, without the
feasibility ceiling, the null-policy score and a classical baseline alongside
it, does not carry enough information to be interpreted. In the case documented
here it led to five rounds of optimisation aimed at a component that was never
the binding constraint --- and, had we stopped at one trainer seed, to a
confident conclusion in whichever direction that seed happened to point.

\section*{Reproducibility}

The prototype this work analyses --- the prediction module, the hexagonal
lattice generator, the placement environment and the training scripts --- is
openly available at
\url{https://github.com/DmytroBabarytskyi/drone-training}, at commit
\texttt{4924613}. The eight experiment scripts, the figure generator and the
JSON record of every run reported here are released alongside it,\footnote{See
the repository's \texttt{README} for the exact location of the experiment
scripts and the archived policy checkpoints.} together with the training
configurations for all three trainer seeds and the corresponding checkpoints.

Three properties of the release matter. Each experiment script imports
the prediction, lattice and environment code from the prototype rather than
reimplementing its formulas, so a validation run tests the implementation that
actually runs. Every Monte Carlo seed is fixed and stated in the sources. And
every policy in Table~\ref{tab:e6} is evaluated over the same episode seeds
through the same environment code path, so the comparison is like-for-like
rather than a comparison of evaluation harnesses. A single entry point,
\texttt{run\_all.py}, regenerates every number and every figure in this paper
in about four minutes on a laptop CPU.

\section*{Declaration on Generative AI}

During the preparation of this work, the author used Claude (Anthropic) in
order to: search and screen the literature; write the experiment scripts
that validate each result by Monte Carlo simulation and the code that produces
Figures~\ref{fig:fig1}--\ref{fig:fig4} from the recorded results; paraphrase
and format the manuscript.

The research programme and its framing, the prototype and simulation
environment that every experiment runs against, the reinforcement-learning
study whose outcomes Section~\ref{sec:learning} reports are the author's own.
Every closed form in this paper is independently checked against Monte Carlo
simulation, and those checks are reported in full. After using this tool, the
author reviewed and edited the content as needed, and takes full responsibility
for the publication's content.


\end{document}